# EfficientSign: An Attention-Enhanced Lightweight Architecture for Indian Sign Language Recognition

Rishabh Gupta, *Member, IEEE*, Shravya R. Nalla, *Member, IEEE*

*Abstract*—How do you build a sign language recognizer that works on a phone? That question drove this work. We built EfficientSign, a lightweight model which takes EfficientNet-B0 and focuses on two attention modules (Squeeze-and-Excitation for channel focus, and a spatial attention layer that focuses on the hand gestures). We tested it against five other approaches on 12,637 images of Indian Sign Language alphabets, all 26 classes, using 5-fold cross-validation. EfficientSign achieves the accuracy of 99.94% (±0.05%), which matches the performance of ResNet18's 99.97% accuracy, but with 62% fewer parameters (4.2M vs 11.2M). We also experimented with feeding deep features (1,280-dimensional vectors pulled from EfficientNet-B0's pooling layer) into classical classifiers. SVM achieved the accuracy of 99.63%, Logistic Regression achieved the accuracy of 99.03% and KNN achieved accuracy of 96.33%. All of these blow past the 92% that SURF-based methods managed on a similar dataset back in 2015. Our results show that attention-enhanced learning model provides an efficient and deployable solution for ISL recognition without requiring a massive model or hand-tuned feature pipelines anymore.

## I. INTRODUCTION

About 6.3 million people in India can't hear. They use Indian Sign Language (ISL) to communicate, and most of the hearing world doesn't understand it. A system that could watch someone sign and translate it in real time would be genuinely useful. Since sign language relies on hand shapes, orientations, and movements to communicate, it is challenging computer vision problem to make an automated recognition system that can work under varying conditions of lighting, skin tone, and background.

Early approaches to ISL recognition relied on handcrafted feature extraction techniques such as Scale-Invariant Feature Transform (SIFT) [1] and Speeded Up Robust Features (SURF) [2], combined with classical machine learning classifiers. While these methods resulted into reasonable accuracy (up to 92% with SVM), they required extensive multi-stage pipelines involving skin masking, edge detection, feature extraction, clustering, and Bag-of-Features construction. Such models are harder to generalize as they are very domain-specific.

The use of deep learning along with transfer learning with pre-trained Convolutional Neural Networks (CNNs), has significantly advanced image classification. Models, which are pre-trained on ImageNet can be fine-tuned on domain-specific datasets, leveraging rich feature hierarchies learned from millions of images. That said, architectures like ResNet [3] contain millions of parameters, which makes it challenging to deploy them on mobile or embedded devices where sign language recognition would be most useful.

In this paper, we propose EfficientSign, a lightweight attention-enhanced architecture that combines EfficientNet-B0 [7] with dual attention mechanisms: Squeeze-and-Excitation (SE) channel attention [8] and Spatial Attention. Our contributions are:

(1) We propose EfficientSign, a novel architecture that integrates SE and Spatial Attention modules with EfficientNet-B0 for ISL recognition, achieving state-of-the-art accuracy with 62% fewer parameters than ResNet18.

(2) We conduct a comprehensive comparative study of six methods (three deep learning, three classical ML) using rigorous 5-fold stratified cross-validation on a 12,637-image ISL dataset.

(3) We demonstrate that deep features (1,280-dimensional vectors from the global average pooling layer) substantially improve classical classifier performance over traditional handcrafted features.

(4) We analyze the trade-off between accuracy and model size (number of parameters), showing EfficientSign is suitable for mobile deployment.

## II. RELATED WORK

Dardas and Georganas [5] implemented real-time gesture recognition using Bag-of-Features (BoF) with SIFT descriptors and SVM classification, demonstrating the viability of visual vocabulary approaches for sign language recognition.

Tharwat et al. [6] applied SIFT features with Linear Discriminant Analysis (LDA) for dimensionality reduction on Arabic Sign Language, achieving 99% accuracy with SVM on 30 sign classes with only 7 training images each.

In our previous work [9], we looked at SURF-based feature extraction using Bag-of-Features and various classifiers, including SVM, KNN, Naive Bayes, and CNN, on a dataset of 4,972 images. We achieved a maximum accuracy of 92% with SVM. This study builds on that by replacing handcrafted features with modern transfer learning and attention mechanisms on a larger dataset.

Wadhawan and Kumar [10] used deep learning-based CNNs to recognize 100 static sign language signs through 35,000 images, demomstrating significant improvements over the traditional methods. Bantupalli and Xie [11] built a real-time system for American Sign Language recognition using CNNs with transfer learning, achieving accuracy of 98.05%.

Our work is different from earlier studies because: (a) we propose a new attention-enhanced architecture designed for lightweight deployment, (b) we provide a direct comparison of six methods under the same experimental conditions with 5-fold cross-validation, and (c) we analyze the trade-off in parameter efficiency explicitly.

## III. DATASET

The dataset consists of 12,637 RGB images of ISL hand gestures representing 26 alphabet classes (A–Z) [12]. Each class contains approximately 486 processed and augmented images captured under varying lighting conditions and backgrounds.

The class distribution is balanced, with each class containing 486 images (class J contains 487). This balance eliminates the need for class weighting or oversampling strategies. For evaluation, we employ 5-fold stratified cross-validation with a fixed random seed (42) to ensure reproducibility, resulting in approximately 10,110 training and 2,527 testing images per fold.

## IV. METHODOLOGY

### A. System Architecture Overview

Figure 1 shows the overall EfficientSign architecture. The pipeline has four stages: (1) Input preprocessing with data augmentation, (2) Feature extraction using the EfficientNet-B0 backbone, (3) Dual attention refinement through SE and Spatial Attention blocks, and (4) Classification using global average pooling and a fully connected layer.

### B. Data Preprocessing

All images are resized to 224×224 pixels to meet the input size required by the pre-trained models. During training, we applied the following augmentations: random horizontal flips, random rotations up to 15 degrees, color jitter (brightness and contrast variations of ±20% and saturation ±10%), and random affine translations up to 10%. We also normalized all images using the ImageNet mean ([0.485, 0.456, 0.406]) and standard deviation ([0.229, 0.224, 0.225]) values. Test images are only resized and normalized.

### C. EfficientSign Architecture

EfficientSign is built on top of EfficientNet-B0 [7], a compound-scaled CNN that scales depth, width, and resolution through a single compound coefficient. Given a set of 224x224 input images, the base EfficientNet-B0 produces 1,280-channel feature maps at 7×7 spatial resolution. The total number of parameters in the EfficientSign model is 4.2M.

Squeeze-and-Excitation (SE) Block [8]: The SE block introduces channel attention by first squeezing global spatial information into a channel descriptor via global average pooling, then learning channel-wise dependencies through two fully connected layers with a reduction ratio r=16. Formally, for input feature map X of shape (C × H × W), the SE block computes: $z = GAP(X)$, $s = \sigma(W_2 \cdot ReLU(W_1 \cdot z))$, and output $= X \odot s$, where $W_1 \in R^{(C/r \times C)}$ and $W_2 \in R^{(C \times C/r)}$. This allows the network to selectively amplify useful channels, such as those encoding hand shape, while downweighting less relevant ones like background texture.

Spatial Attention Block: Unlike the SE block, the spatial attention module focuses on where to attend rather than what. It computes both average-pooled and max-pooled features along the channel axis, concatenates them into a 2-channel map, and applies a 7×7 convolution followed by sigmoid activation to produce a spatial attention map. This directs the model to focus on hand gesture regions while suppressing background areas.

The complete EfficientSign architecture processes input through: (1) EfficientNet-B0 feature extraction (pretrained on ImageNet), (2) SE channel attention, (3) Spatial attention, (4) Global average pooling producing a 1,280-dimensional feature vector, (5) Dropout (p=0.3) for regularization, and (6) A fully connected layer mapping to 26 output classes.

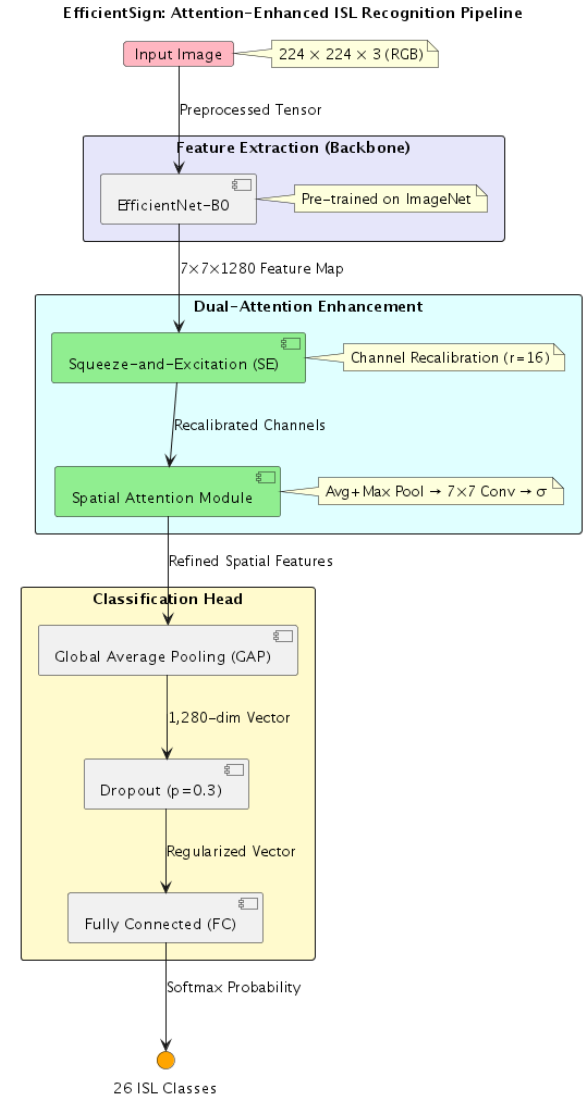

*Fig. 1. EfficientSign System Architecture.*

### D. Baseline Deep Learning Models

ResNet18 [3]: An 18-layer deep residual network using skip connections. We replace the final fully connected layer with a 26-class output. Contains 11.2M parameters.

MobileNetV2 [4]: A lightweight architecture using inverted residuals and linear bottlenecks, designed for mobile applications. We replace the classifier head with a 26-class output. Contains 2.3M parameters.

All deep learning models are fine-tuned for 12 epochs using Adam optimizer with learning rate $1\times10^{-4}$ and cosine annealing learning rate schedule. The best model (by validation accuracy) is saved for final evaluation.

### E. Classical Machine Learning with Deep Features

For classical methods, we use a pre-trained EfficientNet-B0 (with the classification head removed) as a fixed feature extractor. The global average pooling layer outputs a 1,280-dimensional feature vector per image. These deep features replace traditional handcrafted features (e.g., SURF descriptors quantized into Bag-of-Features histograms), providing a richer and more discriminative representation learned from ImageNet's 1.2 million images. Three classifiers are trained on these extracted feature vectors:

(1) Support Vector Machine (SVM): Radial Basis Function (RBF) kernel with regularization parameter C=10 and γ=scale.

(2) K-Nearest Neighbors (KNN): k=5 with uniform weights and Euclidean distance metric.

(3) Logistic Regression: L2 regularization with C=1.0, maximum 1,000 iterations, using the L-BFGS solver.

## V. RESULTS

All experiments were conducted on an Apple MacBook with M-series chip using MPS (Metal Performance Shaders) acceleration. Each model was evaluated using 5-fold stratified cross-validation.

| Method | Params | Mean % | Std % | F1 | F2 | F3 | F4 | F5 |
|---|---|---|---|---|---|---|---|---|
| EfficientSign | 4.2M | 99.94 | ±0.05 | 100.0 | 99.88 | 99.88 | 100.0 | 99.92 |
| ResNet18 | 11.2M | 99.97 | ±0.06 | 100.0 | 100.0 | 100.0 | 100.0 | 99.84 |
| MobileNetV2 | 2.3M | 99.93 | ±0.07 | 99.96 | 100.0 | 99.96 | 99.92 | 99.80 |
| SVM(Deep) | N/A | 99.63 | ±0.11 | 99.45 | 99.72 | 99.56 | 99.72 | 99.68 |
| KNN(Deep) | N/A | 96.33 | ±0.18 | 96.40 | 96.28 | 96.64 | 96.12 | 96.20 |
| LR(Deep) | N/A | 99.03 | ±0.07 | 99.01 | 99.09 | 98.93 | 99.13 | 98.97 |

### A. Deep Learning Results

All three deep learning models achieved near-perfect accuracy. EfficientSign achieved 99.94% (±0.05%), closely matching ResNet18 at 99.97% (±0.06%) while using 62% fewer parameters (4.2M vs 11.2M). MobileNetV2, the lightest model at 2.3M parameters, achieved 99.93% (±0.07%). The low standard deviations across all folds (0.05–0.07%) confirm the robustness and consistency of these results.

### B. Classical ML with Deep Features

Among classical classifiers using 1,280-dimensional deep features from EfficientNet-B0's global average pooling layer, SVM achieved the highest accuracy at 99.63% (±0.11%). Logistic Regression followed at 99.03% (±0.07%). KNN achieved 96.33% (±0.18%), with most errors occurring between visually similar gestures. Notably, even the weakest classical method (KNN at 96.33%) significantly outperforms prior work using handcrafted features (65–92%).

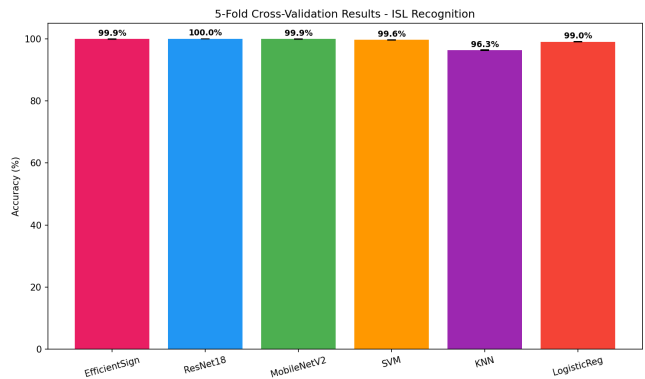

*Fig. 2. 5-Fold Cross-Validation Accuracy Comparison.*

### C. Comparison with Prior Work

| Method | Features | Size | Acc(%) |
|---|---|---|---|
| SVM+SURF+BoF [9] | Handcrafted | 4,972 | 92.00 |
| CNN [9] | Learned | 4,972 | 78.00 |
| KNN+SURF+BoF [9] | Handcrafted | 4,972 | 65.00 |
| Wadhawan [10] | Deep(CNN) | 35,000 | ~99.0 |
| KNN+Deep(Ours) | Deep 1280d | 12,637 | 96.33 |
| SVM+Deep(Ours) | Deep 1280d | 12,637 | 99.63 |
| MobileNetV2 | Deep | 12,637 | 99.93 |
| EfficientSign | Deep+Attn | 12,637 | 99.94 |
| ResNet18 | Deep | 12,637 | 99.97 |

## VI. DISCUSSION

One of the biggest advantages of transfer learning is that it removes the need for manual feature engineering altogether. In our earlier work [9], we had to go through multiple stages of applying skin masks, running Canny edge detection, extracting SURF features, clustering them into 150 visual words with K-means, and then building Bag-of-Features histograms. EfficientSign eliminates all of that by learning directly from raw images in an end-to-end fashion, improving accuracy from 92% up to 99.94%.

We also observed that feeding deep features into traditional classifiers improves their performance. When we replaced old SURF+BoF representations for 1,280-dimensional vectors from EfficientNet-B0's global average pooling layer, SVM accuracy increased from 92% to 99.63% on a larger dataset. This implies that the limitation in older methods wasn't the choice of classifier, but it was the quality of the features being fed into it.

What stands out most about this research is how the attention models match the performance of larger models with fewer parameters. EfficientSign hits 99.94% accuracy using just 4.2M parameters, while ResNet18 needs 11.2M to reach 99.97%—that's 62% more parameters for a negligible 0.03 point edge. The SE block helps the network zero in on channels that capture hand shape information, and the Spatial Attention module keeps the focus on the hand itself even when backgrounds change. Together, these make EfficientSign a practical choice for running on phones or other resource-constrained hardware.

| Model | Params | Acc(%) | Rel. Size |
|---|---|---|---|
| MobileNetV2 | 2.3M | 99.93 | 1.0× |
| EfficientSign | 4.2M | 99.94 | 1.8× |
| ResNet18 | 11.2M | 99.97 | 4.9× |

We should be clear about our limitations. The results show 99.94% accuracy across all deep learning models, but this dataset contains clean backgrounds and consistent lighting. Real world sign language happens in homes, in public places, and in poorly lit environments. We haven't tested how the model handles partial hand occlusion, motion blur, or live video input. The dataset also only covers static alphabet gestures (letters like J and Z, which require motion, are excluded), and real ISL conversations go beyond spelling individual letters. They involve whole words and sentences. This work is a starting point, not a finished product.

## VII. CONCLUSION

In this paper, we introduced EfficientSign, a lightweight model designed to recognize Indian Sign Language alphabets using attention mechanisms. It is built on an EfficientNet-B0 backbone combined with Squeeze-and-Excitation channel attention with Spatial Attention modules, it achieved 99.94% mean accuracy through 5-fold cross-validation on 12,637 ISL images. This matches the performance of ResNet18 while using 62% fewer parameters. In our study of six methods, results show that transfer learning combined with attention modules offers a strong, efficient, and deployable solution for ISL recognition, significantly surpassing previous handcrafted feature methods. Moving forward, we aim to expand this work to include dynamic gesture recognition, real-time video-based ISL translation, and on-device deployment for mobile platforms.

## REFERENCES

[1] D. G. Lowe, "Distinctive image features from scale-invariant keypoints," Int. J. Comput. Vis., vol. 60, no. 2, pp. 91-110, 2004.
[2] H. Bay, T. Tuytelaars, and L. Van Gool, "SURF: Speeded up robust features," in Computer Vision – ECCV 2006, pp. 404-417, Springer, 2006.
[3] K. He, X. Zhang, S. Ren, and J. Sun, "Deep residual learning for image recognition," in Proc. IEEE CVPR, pp. 770-778, 2016.
[4] M. Sandler et al., "MobileNetV2: Inverted residuals and linear bottlenecks," in Proc. IEEE CVPR, pp. 4510-4520, 2018.
[5] N. H. Dardas and N. D. Georganas, "Real-time hand gesture detection and recognition using bag-of-features and SVM," IEEE Trans. Instrum. Meas., vol. 60, no. 11, pp. 3592-3607, 2011.
[6] A. Tharwat et al., "SIFT-based Arabic sign language recognition system," in Afro-European Conf. for Industrial Advancement, pp. 359-370, 2015.
[7] M. Tan and Q. V. Le, "EfficientNet: Rethinking model scaling for CNNs," in Proc. ICML, pp. 6105-6114, 2019.
[8] J. Hu, L. Shen, and G. Sun, "Squeeze-and-Excitation Networks," in Proc. IEEE CVPR, pp. 7132-7141, 2018.
[9] R. Gupta, S. Agarwal, S. Kumar, and V. Behl, "Indian Sign Language Recognition Using Image Processing," B.Tech Project Report, DTU, 2015.
[10] A. Wadhawan and P. Kumar, "Deep learning-based sign language recognition for static signs," Neural Comput. & Applic., vol. 32, pp. 7957-7968, 2020.
[11] K. Bantupalli and Y. Xie, "ASL recognition using deep learning and computer vision," in Proc. IEEE Int. Conf. Big Data, pp. 4896-4899, 2018.
[12] A. Tasnim, "Indian Sign Language Dataset," GitHub, 2020. https://github.com/ayeshatasnim-h/Indian-Sign-Language-dataset